\documentclass[runningheads]{llncs}
\usepackage[T1]{fontenc}
\usepackage{xurl}
\usepackage{hyperref}
\usepackage{tabularx}
\usepackage{amsmath}
\usepackage{amssymb}

\usepackage{array}
\usepackage{graphicx}
\graphicspath{
	{figures_main/}
	{figures_supplementary/}
}
\begin{document}
\title{Longitudinal Learner Participation Profiling in Online Learning: Heterogeneous Consensus Clustering under Unequal Observation Windows}
\titlerunning{Contextual Learner--Course Participation Profiles}
%
%
\author{
David Otoo-Arthur\inst{1}\orcidID{0000-0003-2057-088X} \and
Terence L. van Zyl\inst{2}\orcidID{0000-0003-4281-630X}
}
\authorrunning{D. Otoo-Arthur and T. L. van Zyl}
\institute{
University of the Witwatersrand, Johannesburg, South Africa\\
\email{david.otoo-arthur1@students.wits.ac.za}
\and
University of Johannesburg, Johannesburg, South Africa\\
\email{tvanzyl@uj.ac.za}
}
\maketitle              

\begin{abstract}

Learning management system traces are clustered to identify participation profiles, but unequal observation windows may cause profiles to reflect duration rather than participation alone. Using 16,448 complete learner--course sequences from 4,599 learners across 152 courses at a South African public university, we evaluated eight participation measures. All remained related to observation-window length even when expressed as rates or shares. Course-grouped adjustment reduced these relationships. Five retained measures were evaluated using K-means, Gaussian mixtures, Ward agglomerative clustering and K-medoids, with standardised measures and principal-component summaries. Evidence supported a broad two-profile grouping and a more detailed adjusted four-profile consensus: Gradually declining, Early-concentrated, Cyclical variable and Sustained active. Profile membership accounted for 2.87\% of the variation in duration in the primary consensus, versus 38.18\% using original measures. Mean assignment support was 0.739; 8.35\% of assignments fell below 0.50, with greatest uncertainty in Sustained active. Two week-by-week sequence analyses agreed strongly with each other but only modestly with the four-profile consensus, indicating that the profiles combined multiple aspects of participation rather than distinct sequence patterns. Profiling should therefore assess duration effects, compare clustering methods and participation summaries, and report assignment consistency.

	
\end{abstract}

\section{Introduction}
\label{sec1}

Learning management systems (LMSs) are widely used to support teaching, assessment and course administration in higher education \cite{turnbull2020}. As learners access resources, submit assessments, participate in forums and use other platform tools, they generate time-stamped traces of their interactions with the digital course environment. These records provide a scalable basis for examining how recorded participation develops over time \cite{henrie2015,saqr2023achievement}. Their interpretation nevertheless requires care. LMS traces capture observable platform interaction rather than learning or engagement in its broader psychological sense. Engagement encompasses behavioural, cognitive and emotional involvement \cite{fredricks2004,boekaerts2016}, whereas platform records primarily represent its observable behavioural component. Learners may study downloaded materials offline, use channels outside the LMS or generate recorded events without equivalent cognitive investment \cite{henrie2015,wise2015}.

Recorded LMS participation is commonly summarised using event totals, logins, resource views, downloads or active days \cite{henrie2015,viberg2018}. Although these measures describe how much learners use the platform, they provide limited information about how participation develops over time. Similar event volumes can arise from sustained participation, activity concentrated in the early weeks, repeated interruption and return, or gradual decline. For this reason, learning analytics studies increasingly examine weekly participation patterns, the order in which those patterns occur and longer-term changes, rather than activity volume alone \cite{jovanovic2017,barthakur2021,papageorgiou2025,heikkinen2025}. These approaches consistently demonstrate that learner participation is heterogeneous and dynamic. These approaches examine participation at different levels. Weekly models describe short-term changes, sequence models retain the order of weekly participation patterns, and longer-term trajectories show how participation develops across courses or years.

Courses may differ in the number of weeks for which participation can be observed. We refer to this period as the \emph{observation window}. The length of this window affects the number of changes and the periods of activity or inactivity that can be observed \cite{studer2016,ritschard2023}. Expressing activity per week, which is a rate, or as a proportion of the observed weeks, which is a share, can improve comparability. However, this does not completely remove the effect of course length. For example, one active week represents one-quarter of a four-week observation window but only one-twelfth of a twelve-week observation window. Consequently, records may be grouped partly by the number of observed weeks rather than by participation patterns alone.

Clustering introduces a second source of uncertainty. Different clustering methods and different ways of summarising participation can produce different but equally reasonable groupings. Consensus clustering reduces reliance on any one result by combining evidence from several groupings \cite{li2021ensemble,strehl2002}. However, it cannot correct measures that are still affected by the observation-window length, identify a single unquestionably correct number of profiles, or guarantee that every case is consistently assigned.

Although previous studies have examined how learner participation changes over time, how weekly patterns develop and how results from different grouping methods can be combined, these areas have usually been studied separately. Few studies have examined three issues together: whether the number of observed weeks affects participation measures, whether similar profiles emerge when different measures and grouping methods are used, and how consistently each record is assigned to the same profile. It therefore remains unclear whether the resulting profiles mainly reflect learners’ participation over time, differences in the number of weeks observed or choices made during the analysis.

Our study addresses these problems using longitudinal Moodle records from a public university in South Africa. African higher-education institutions remain comparatively underrepresented in international learning-analytics research \cite{prinsloo2022}, despite expanding use of LMS-supported and blended learning and continuing differences in access, infrastructure and institutional capacity \cite{audigitaleducation2023,audigitaltransformation2020}. This setting matters because LMS activity is shaped by particular course, institutional and access conditions. However, the same problem applies in other settings. A study comparing participation across several courses may conflate differences in course length with differences in learner participation if the number of observed weeks is not examined explicitly.

Drawing on methods from artificial intelligence, statistical learning and learning analytics, we develop a consensus-clustering framework that accounts for unequal observation windows. The framework first checks whether participation measures are affected by the number of observed weeks and adjusts them where necessary. It then compares profiles across different summaries of participation and clustering methods and assesses how consistently each learner--course record is assigned, distinguishing well-supported assignments from less certain ones. Accordingly, the profiles are interpreted as patterns of participation within particular courses rather than as fixed labels for learners.

\subsection{Contributions}
\label{sec:contributions}

Our study makes three contributions. First, we show that rate- and share-based LMS measures can remain associated with observation-window length, which can reduce comparability across courses. Second, we develop a consensus-clustering framework that accounts for observation-window length, combines results from different participation summaries and grouping methods, and assesses the consistency with which each record is assigned. Third, we identify four interpretable learner--course participation profiles and examine their stability through individual assignment consistency and comparisons across different summaries and week-by-week sequences. We interpret the profiles as context-specific participation patterns rather than fixed learner types.

\section{Related Work and Conceptual Background}
\label{sec:related_work}

\subsection{From Aggregate Activity to Temporal Participation}
\label{sec:temporal_participation}

A substantial body of learning analytics research has used aggregate LMS indicators, including event counts, login frequency, active days and resource views. Reviews by Henrie et al. \cite{henrie2015} and Viberg et al. \cite{viberg2018} show that such indicators are widely used to examine achievement, persistence and engagement. Although convenient for analysis, combining activity into totals removes information about when the activity occurred and in what order. Similar totals may therefore represent sustained participation, isolated activity or declining participation.

Temporal approaches have addressed this limitation at different analytical levels. Person-oriented research focuses on recurring patterns of behaviour among learners rather than average relationships between separate measures \cite{bergman1997}. Within individual courses, Jovanovi\'c et al. \cite{jovanovic2017} identified learning strategies from combinations of course activities. Papageorgiou et al. \cite{papageorgiou2025} subsequently identified active and inactive weekly effort--time profiles and examined how learners moved between them. These studies show that participation varies within the same course, although weekly profiles describe short-term patterns rather than participation over the complete course.

Sequence-based methods preserve temporal order more directly. Heikkinen et al. \cite{heikkinen2025}, for example, used multichannel sequence analysis, which tracks multiple forms of behaviour over time, to examine the relationship between behavioural engagement and self-regulation. These methods retain the order of the recorded participation categories, but their findings depend on how those categories are defined and how differences between sequences are calculated \cite{studer2016}. Longer-term studies have instead examined patterns across courses or programmes, including programme-level learning strategies \cite{barthakur2021} and trajectories of engagement, achievement and learning tactics \cite{saqr2021,saqr2023achievement,saqr2023tactics}.

Course-level profiling provides another perspective by summarising several characteristics of each learner or enrolment. Zamecnik et al. \cite{zamecnik2022} combined trace, demographic, and motivational evidence to examine profiles of online learners, whereas Roski et al. \cite{roski2024} clustered patterns of digital learning resource use. These approaches compare participation across several measures but often overlook how activity unfolds within a course. Including measures of change over time can provide a more complete learner--course profile. However, research has paid less attention to whether these measures remain comparable across courses with different observation windows.

\subsection{Clustering, Ensemble Evidence and Uncertainty}
\label{sec:clustering_literature}

Clustering is widely used to identify groups with similar behaviour when those groups are not known in advance. However, the resulting groups depend on how the data are summarised and which grouping method is used. Jain \cite{jain2010} describes several families of clustering methods, including partitioning, hierarchical and model-based approaches, which define similarity and group structure differently. Studies using educational data have likewise shown that different grouping methods can produce groups that differ in clarity and stability \cite{navarro2018}.

One way to reduce this dependence is ensemble clustering, which combines several grouping results into a one overall solution. Strehl and Ghosh \cite{strehl2002} formalised this approach for combining results generated using different clustering methods and different sets of measures. In educational research, Li et al. \cite{li2021ensemble} applied ensemble clustering to student behavioural data and reported that the combined result was less dependent on any one grouping. These studies support using evidence from multiple methods and measures rather than relying on a single solution.

However, consistent clustering results do not necessarily mean that the groups are educationally meaningful. Hennig \cite{hennig2007} showed that stable clusters can arise because a method tends to reproduce the same grouping rather than because the groups represent meaningful patterns. Similarly, several methods may agree because they all use data containing the same unwanted pattern. If participation measures remain related to course duration, agreement among the methods cannot by itself show that the profiles represent participation rather than observation-window length.

Consensus methods can also show how consistently individual cases are assigned, rather than only whether the overall grouping is stable. Monti et al. \cite{monti2003} demonstrated that examining how often cases are placed together across repeated analyses can reveal stable and unstable assignments. In educational profiling, this distinguishes cases that are repeatedly assigned to the same profile from cases that move between profiles when analytical choices change. Reporting only one final cluster label hides this uncertainty and may encourage readers to treat a grouping that depends on the data and method as a fixed learner characteristic.

Choosing the number of clusters presents a related difficulty. The silhouette coefficient assesses whether cases are closer to their own cluster than to neighbouring clusters \cite{rousseeuw1987}. The Calinski--Harabasz index compares separation between clusters with variation within them \cite{calinski1974}, while the Davies--Bouldin index assesses how similar the clusters are to one another \cite{davies1979}. For mixture models, the Akaike Information Criterion (AIC) and Schwarz Bayesian Information Criterion (BIC) compare model fit while penalising unnecessarily complicated solutions \cite{akaike1974,schwarz1978}. Because these criteria assess different aspects of a clustering result, they may not favour the same number of clusters. Akhanli and Hennig \cite{akhanli2020} therefore argue that cluster validation should reflect the purpose and desired characteristics of the grouping rather than depend on one index.

\subsection{Duration Comparability}
\label{sec:duration_comparability}

Sequence length is an important characteristic of data collected over time. Ritschard \cite{ritschard2023} reviewed measures describing sequence length, changes between participation categories, uninterrupted periods in one category, known as \emph{spells}, and broader measures of how varied a sequence is, termed \emph{complexity} and \emph{turbulence}. Studer and Ritschard \cite{studer2016} showed that sequences may differ in when participation occurs, the order in which participation categories appear and how long each uninterrupted period lasts. When observation windows differ, longer sequences provide more opportunities to observe changes between categories, returns to activity and long periods in the same category.

Converting counts into rates or shares can make values easier to compare, but it does not fully remove the effect of observing courses for different numbers of weeks. A single change has greater proportional influence in a short sequence, whereas longer sequences allow longer periods of inactivity and more weeks without a change. Rescaling measures to a common scale, known as standardisation, changes their numerical values but may not remove their relationship with observation length. Similarly, reducing many measures to a smaller set, known as dimensionality reduction, may preserve differences related to course length. Comparability across observation windows should therefore be tested using the data rather than assumed from how a measure is calculated.

\subsection{Research Questions}
\label{sec:research_questions}

The issues identified in the literature motivate the following research questions:

\begin{description}
	
    \item[RQ1.] How strongly are participation measures related to the number of observed course weeks, and how effectively does duration adjustment reduce these relationships?
    
    \item[RQ2.] What learner--course participation profiles emerge after duration adjustment, and how consistently are records assigned to them?
    
    \item[RQ3.] How stable are these profiles across grouping methods, participation summaries and numbers of clusters, and how closely do they match profiles based on week-by-week sequences?
	
\end{description}

\section{Methodology}
\label{sec:methodology}

\subsection{Data, Cohort and Indicators}

To address these research questions, we used a retrospective longitudinal design and anonymised Moodle records from a public South African university. The data covered February 2013--August 2019 and comprised 1,172,894 learner--course--week observations from 5,020 learners, 179 courses and 30,610 learner--course units. Ethical and institutional approval was obtained for secondary analysis. Because the records did not capture offline participation, the analysis was restricted to observable Moodle activity.

Course-specific endpoints were defined to accommodate systematic one-week differences in final-week availability. Ten courses with only one observed week were excluded because longitudinal change could not be estimated. Validation confirmed that every retained sequence began at Week~1, reached its course endpoint and contained neither duplicated nor internally missing weeks. The resulting longitudinal dataset contained 169 courses and 29,736 learner-course units. A week was classified as active when at least one Moodle event was recorded. Participation intensity was represented using the log-transformed event count, calculated as $\log(1+\text{event count})$. Units with no active week ($n=6,817$) or one active week ($n=6,471$) were retained as descriptive categories but excluded from dynamic clustering. The profiling cohort, therefore, comprised 16,448 sequences, 4,599 learners, 152 courses and 587,276 weekly observations.

Eight indicators were calculated to describe participation frequency and how participation changed over time \cite{studer2016,ritschard2023}: active-week rate; transition rate, representing changes between active and inactive weeks; reactivation rate, representing returns to activity after inactivity; longest active- and inactive-run shares, representing the longest consecutive active and inactive periods; mean absolute week-to-week change in the log-transformed event count; within-sequence standard deviation of the log-transformed event count; and the share of total events occurring in the most active week. Transition and reactivation rates were divided by the $T_{ic}-1$ possible week-to-week changes, while the longest-run measures were divided by sequence duration $T_{ic}$. Completeness, finiteness and logical bounds were checked before analysis.

\subsection{Duration Adjustment and Representations}
For learner $i$ in course $c$, we denote the number of observed weeks by $T_{ic}$ and the observed value of participation indicator $j$ by $X_{icj}$. To assess whether the indicators were affected by observation-window length, each indicator was correlated with $T_{ic}$ using Pearson and Spearman coefficients and with $\log(T_{ic})$ using Pearson coefficients. Means were also compared across duration groups. Because several exact-duration levels were represented by few courses, expected indicator values were estimated with a spline--ridge model using five-fold course-grouped cross-fitting. Thus, the model used to predict a sequence never included observations from its course. For indicator $X_j$, the duration-relative value was calculated by subtracting the value expected for its observation-window length from its observed value:

\begin{equation}
	X^{\mathrm{rel}}_{icj}
	=
	X_{icj}-\widehat{f}^{(-c)}_j(T_{ic}),
	\label{eq:duration_residual}
\end{equation}

where $\widehat{f}^{(-c)}_j$ was estimated without the held-out course group. Duration correlations were recalculated after adjustment.

Pearson and Spearman correlations, together with variance inflation factors, were used to identify indicators that provided similar information. This process retained five indicators: active-week rate, reactivation rate, longest inactive-run share, mean absolute log change and maximum-week event share. These indicators were prepared for clustering in four forms: standardised original indicators; standardised duration-relative indicators; three principal components from the original indicators, rescaled to have equal variance; and three equivalent components from the duration-relative indicators.

\subsection{Clustering, Consensus and Sensitivity Analysis}

K-means was evaluated for $k=2,\ldots,8$ on all four representations. Gaussian-mixture, connectivity-constrained Ward agglomerative and K-medoids models were evaluated for $k=2,\ldots,6$, thereby allowing for different cluster shapes and different ways of assigning cases \cite{jain2010,navarro2018}. Candidate solutions were evaluated using standard measures of cluster separation and compactness—the silhouette, Calinski--Harabasz and Davies--Bouldin indices—together with minimum cluster size and, where applicable, inertia and uncertainty in profile membership \cite{rousseeuw1987,calinski1974,davies1979,akhanli2020}. Gaussian-mixture models were also evaluated using the Akaike Information Criterion (AIC) and Bayesian Information Criterion (BIC), which favour models that fit the data well without becoming unnecessarily complex; lower values indicate a better balance \cite{akaike1974,schwarz1978}. K-medoids used FasterPAM on a reproducible sample of 3,000 sequences; remaining sequences were assigned to their nearest medoid \cite{schubert2021}. Adjusted Rand index (ARI) quantified agreement across algorithms and representations \cite{hubert1985}.

Both a broad $k=2$ structure and a finer $k=4$ structure were carried into consensus analysis. For each resolution, cluster labels were aligned and converted into binary indicators of cluster membership. The full heterogeneous consensus combined 16 clustering results obtained from four algorithms and four representations; the original-only and duration-relative variants each combined eight results. This design reduced dependence on a single clustering specification \cite{strehl2002,monti2003,li2021ensemble}. Assignment support was the proportion of contributing clustering results that agreed with each record’s final profile. Eta-squared measured the relationship between profile membership and observation-window length. For $k=4$, the consensus solution with the weakest relationship was selected for primary interpretation. The stability of the consensus was assessed by removing one algorithm at a time and comparing results across the different indicator forms.

Profiles were interpreted from raw indicator means, standardised duration-relative centroids and high-support cores. Support of at least 75\% was classified as high, 50--74.9\% as moderate and below 50\% as low. Activity trajectories were calculated at 20 equally spaced points along the relative course of progress, with each point mapped to the nearest observed week.

Finally, two sequence-based sensitivity analyses tested whether the profiles were reproduced from temporal shape. At each of the 20 course-progress points, participation was coded as active if at least one Moodle event was recorded and as inactive if no event was recorded. The resulting sequences were compared using Hamming distance, which counts the points at which they differ. A multichannel representation combined 20 activity-state and 20 standardised log-intensity points using Euclidean distance. FasterPAM models for $k=2,\ldots,6$ were fitted to a reproducible sample of 4,000 sequences and extended by nearest-medoid assignment. Agreement with the primary consensus was evaluated using ARI and aligned exact agreement for all sequences and high-support cores. These comparisons were treated as sensitivity analyses because indicator- and distance-based clustering encode different behavioural information.

\section{Results}
\label{sec:results}

\subsection{Duration Sensitivity and Adjustment (RQ1)}

Excluding the ten one-week courses removed 5.59\% of courses but only 2.86\% of learner--course units and 0.075\% of weekly observations. Of the 29,736 eligible sequences, 55.31\% formed the profiling cohort; the remainder had no active week (22.93\%) or one active week (21.76\%).

\begin{figure}[tbp]
	\centering
	\includegraphics[width=0.90\linewidth]
	{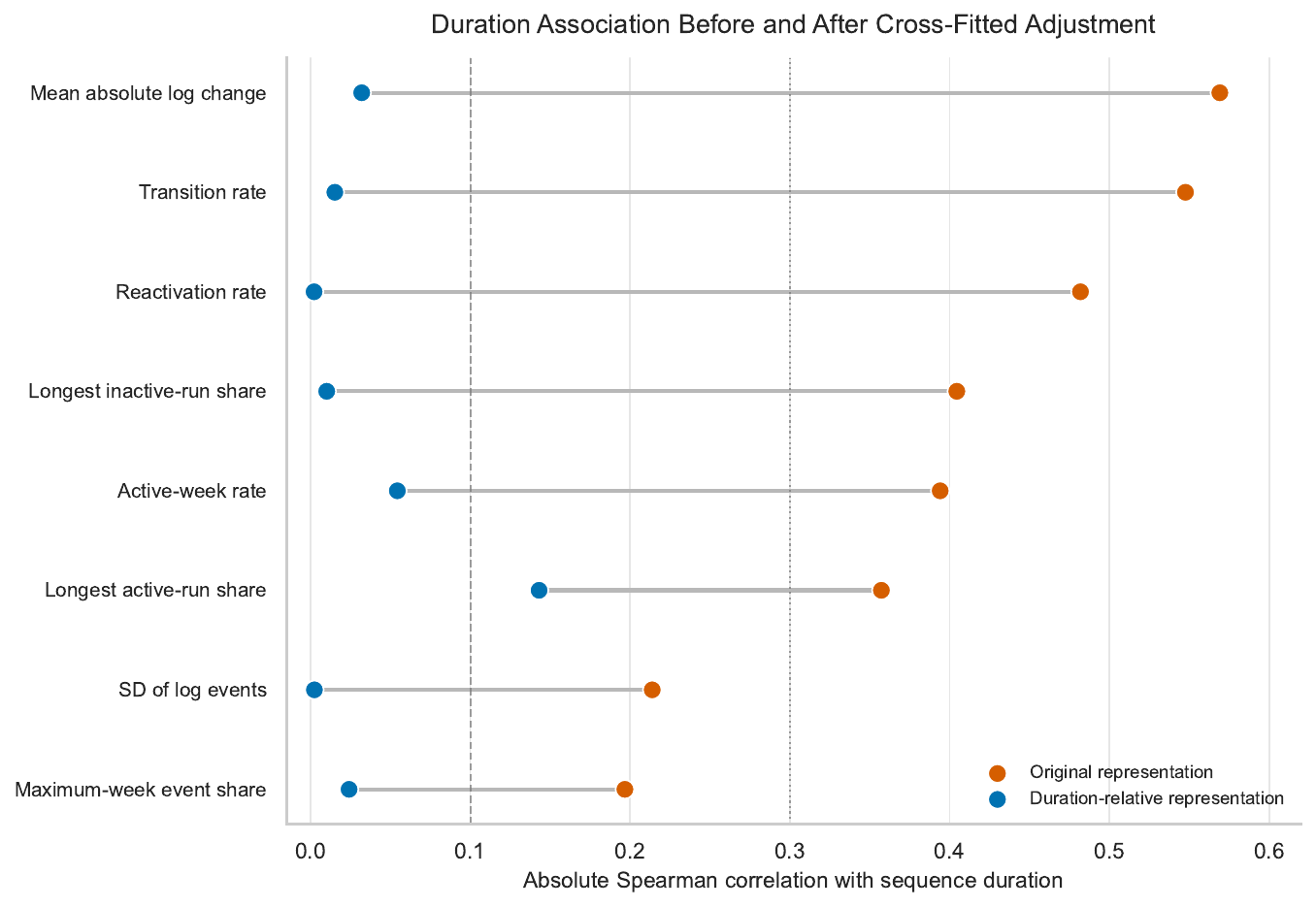}
	\caption{Course-grouped adjustment reduced the absolute Spearman correlations with observation-window length for all eight measures, from 0.197--0.569 before adjustment to 0.002--0.143 afterwards. Seven adjusted correlations were below 0.06.}
	\label{fig:duration_adjustment}
\end{figure}

All eight candidate measures remained associated with observation-window length, including those expressed as rates or shares. Before adjustment, absolute Spearman correlations ranged from 0.197 to 0.569, with six above 0.30. Course-grouped adjustment reduced these correlations to 0.002--0.143, with seven below 0.06 (Fig.~\ref{fig:duration_adjustment}). After adjustment, Pearson correlations were also near zero, ranging from $-0.025$ to 0.012. The adjusted forms of the measures were therefore used for primary profiling.

The eight measures also contained overlapping information. Transition and reactivation rates correlated at approximately 0.97, while the correlation between active-week rate and longest active-run share exceeded 0.89. Selecting five measures reduced the maximum VIF from 16.73 to 6.53 for the original measures and from 15.49 to 6.05 for the duration-relative measures.

\subsection{Consensus Profiles and Assignment Support (RQ2)}

The evaluation criteria did not point to a single number of profiles. Standardised results generally favoured $k=2$, while PCA-based results from the original measures favoured $k=4$. For Gaussian-mixture models, AIC and BIC continued to favour more groups through $k=6$, although uncertainty in profile membership increased. Mean agreement across grouping methods, measured by ARI, was highest at $k=2$ (0.541) and next highest at $k=4$ (0.477). We therefore retained $k=2$ as the broad grouping and $k=4$ as the more detailed grouping.

The broad $k=2$ consensus was balanced (53.22\% versus 46.78\%) and strongly supported (mean support $=0.867$). However, profile membership accounted for 31.85\% of the variation in observation-window length in the full $k=4$ consensus and 38.18\% in the original-only consensus, compared with only 2.87\% in the duration-relative consensus. The range of mean observation-window lengths across profiles also decreased from 23.35 to 7.27 weeks. The duration-relative $k=4$ consensus was therefore selected as the primary detailed solution.

\begin{figure}[tbp]
	\centering
	\includegraphics[width=0.90\linewidth]
	{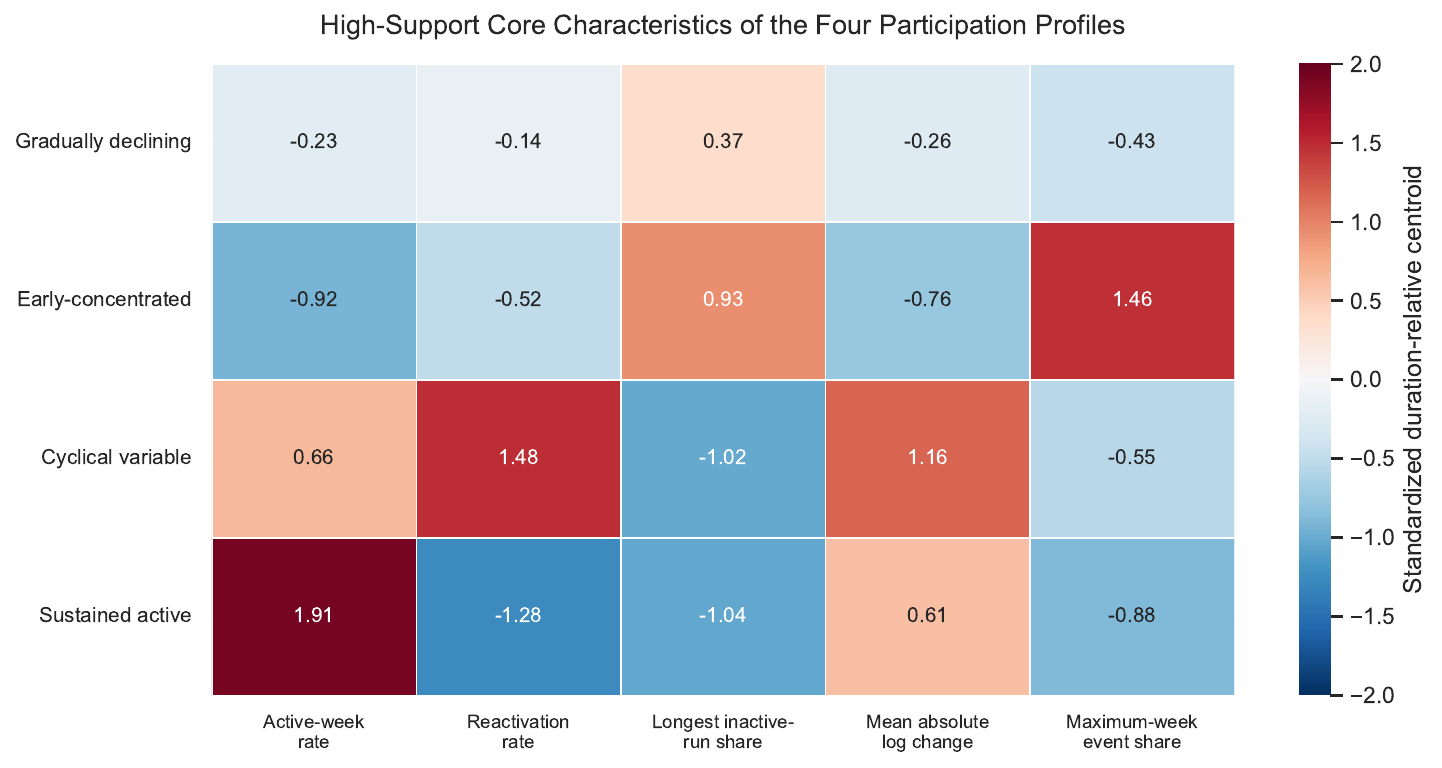}
	\caption{Average standardised duration-adjusted indicator values for cases with high assignment support in each profile. Positive values indicate characteristics above, and negative values below, those expected for records with similar observation-window lengths.}
	\label{fig:profile_heatmap}
\end{figure}

The four profiles were Gradually declining ($n=4,873$), Early-concentrated ($n=5,191$), Cyclical variable ($n=3,594$) and Sustained active ($n=2,790$). Table~\ref{tab:profiles} reports their indicator means and assignment support, while Fig.~\ref{fig:profile_heatmap} shows the adjusted indicator values for cases with high assignment support.

\begin{table}[tbp]
	\centering
	\caption{Characteristics and support of the primary participation profiles}
	\label{tab:profiles}
	\fontsize{7.1}{8.0}\selectfont
	\setlength{\tabcolsep}{2.0pt}
	\begin{tabular}{lrrrrrrrr}
		\hline
		\textbf{Profile} & \textbf{$n$} & \textbf{\%} &
		\textbf{AWR} & \textbf{RR} & \textbf{LIRS} &
		\textbf{MALC} & \textbf{MES} & \textbf{Support} \\
		\hline
		Gradually declining & 4,873 & 29.63 & 0.244 & 0.083 &
		0.539 & 0.532 & 0.349 & 0.736 \\
		Early-concentrated & 5,191 & 31.56 & 0.110 & 0.060 &
		0.677 & 0.319 & 0.678 & 0.788 \\
		Cyclical variable & 3,594 & 21.85 & 0.455 & 0.164 &
		0.249 & 1.087 & 0.304 & 0.743 \\
		Sustained active & 2,790 & 16.96 & 0.622 & 0.061 &
		0.250 & 0.946 & 0.231 & 0.649 \\
		\hline
		\multicolumn{9}{p{0.96\linewidth}}{\textit{Note.}
			AWR = active-week rate; RR = reactivation rate;
			LIRS = longest inactive-run share; MALC = mean absolute log change;
			MES = maximum-week event share. Support is the mean proportion of eight clustering results that agreed with the final profile assigned to each record.}
	\end{tabular}
\end{table}

\begin{figure}[tbp]
	\centering
	\includegraphics[width=0.90\linewidth]
	{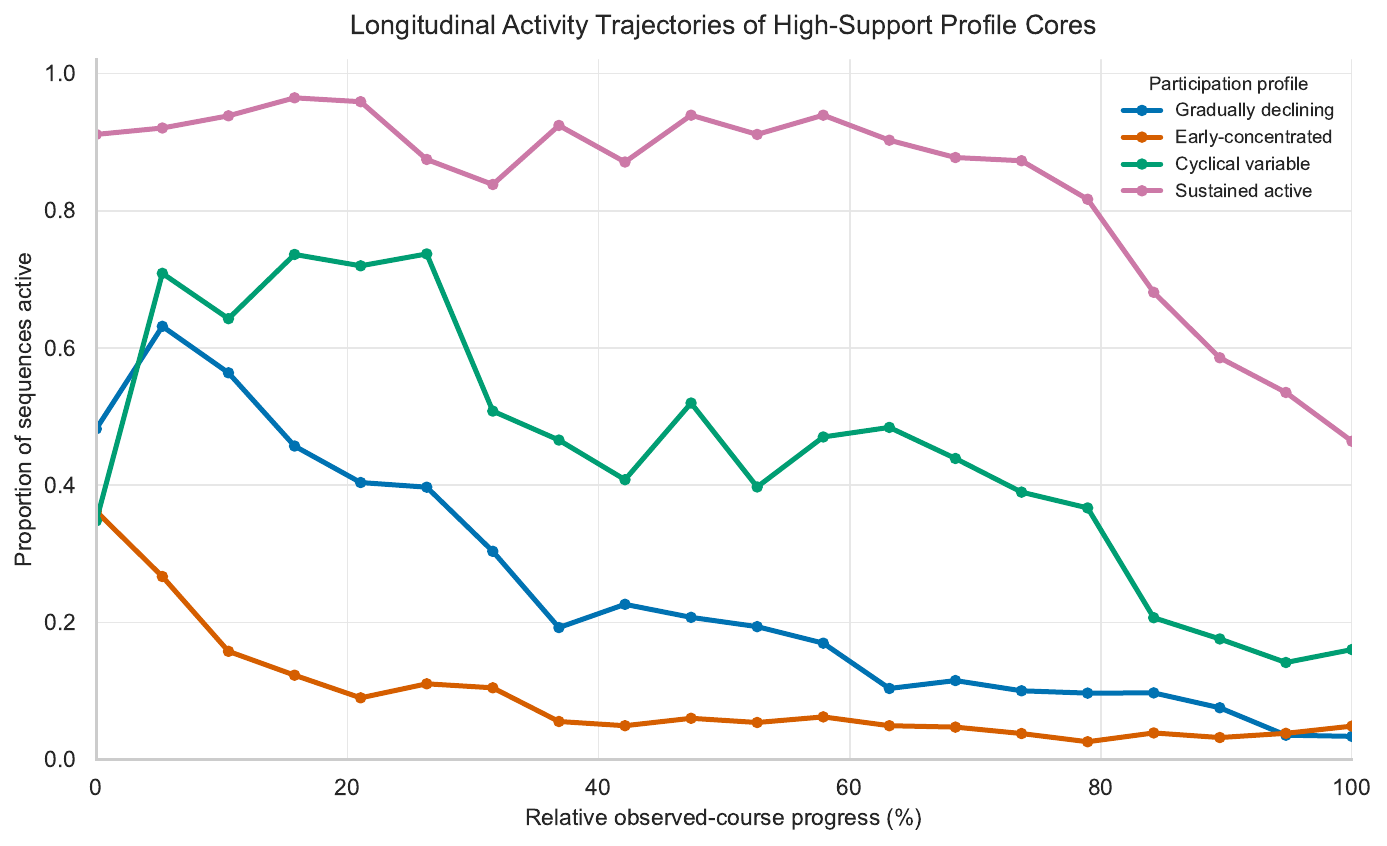}
	\caption{The high-support trajectories distinguished four participation profiles: Early-concentrated activity declined rapidly after course entry; Gradually declining activity peaked early and then decreased progressively; Cyclical variable activity fluctuated throughout the course; and Sustained active participation remained high until a late-course decline.}
	\label{fig:profile_trajectories}
\end{figure}

Gradually declining participation increased briefly at the beginning and then decreased progressively. Early-concentrated participation occurred mainly near the beginning, with the lowest active-week rate, longest inactivity and greatest concentration of events in one week. Cyclical variable participation showed repeated interruptions and returns, with the highest reactivation and week-to-week variation. Sustained active participation remained relatively high through most of the course but declined near the end (Fig.~\ref{fig:profile_trajectories}).

Mean assignment support was 0.739: 61.36\% of assignments received support of at least 0.75 and 8.35\% fell below 0.50. Early-concentrated had the highest mean support (0.788), while Cyclical variable had no assignments below 0.50. Sustained active was the least certain, with 26.02\% of its assignments below 0.50. Interpretation therefore focused on the 10,093 sequences with support of at least 0.75.

\subsection{Robustness and Sequence-Based Sensitivity (RQ3)}

Removing Gaussian mixtures or agglomerative clustering produced results similar to the primary consensus (ARI $=0.897$ and 0.869), whereas removing K-medoids or K-means produced larger changes (ARI $=0.463$ and 0.458). Results based directly on the standardised duration-relative measures agreed more strongly with the primary solution than the PCA-based results (ARI $=0.805$ versus 0.496). Thus, the profiles could be interpreted, but the detailed grouping depended partly on K-means and K-medoids and on using the measures directly rather than their principal-component summaries.

Sequence clustering favoured the broader two-profile grouping. For the active--inactive sequences, the silhouette score was 0.416 at $k=2$ and 0.303 at $k=4$. The four-profile sequence result showed modest agreement with the primary consensus (ARI $=0.280$ overall and 0.354 for cases with high assignment support). Multichannel agreement was similar (ARI $=0.235$ overall and 0.334 for high-support cases), although the active--inactive and multichannel clustering results agreed strongly with each other (ARI $=0.710$). The sequence analyses therefore found a consistent broad week-by-week pattern but reproduced only part of the four profiles based on the summary measures.

\section{Discussion}
\label{sec:discussion}

The main methodological finding was that converting participation measures into rates or shares was not enough to make records from shorter and longer courses comparable. In a shorter record, one active week or one change has a greater effect on the measures than it does in a longer record. Before adjustment, profile membership accounted for 38.18\% of the variation in observation-window length; after adjustment, this fell to 2.87\%. The adjustment for each course was estimated using data from other courses. It therefore greatly reduced the influence of observation-window length, although differences in course design and context may still affect the profiles.

The four profiles revealed differences that overall activity totals would not show. Gradually declining and Early-concentrated participation both involved limited activity but differed in when it occurred and how concentrated it was. Cyclical variable and Sustained active participation both involved substantial activity but differed in the frequency of interruptions and returns. This supports previous evidence that patterns over time provide information beyond event totals \cite{jovanovic2017,papageorgiou2025,heikkinen2025}. The labels describe only recorded Moodle participation. Inactivity does not establish psychological disengagement or its cause, and the profiles describe participation within particular courses rather than fixed learner types.

The four-profile solution was not the only reasonable grouping. Several evaluation criteria supported the broader two-profile result, and the more detailed solution depended particularly on K-means and K-medoids. Assignment support showed that many records were consistently placed in the same profile, while others had less certain assignments, particularly within Sustained active participation. Assignment support should therefore accompany profile labels, and low-support assignments should not be used for automated intervention.

The modest agreement with sequence-based clustering further limits the interpretation. The active--inactive and multichannel methods compared participation at corresponding course points, while the five selected measures summarised activity frequency, returns after inactivity, length of inactivity, week-to-week variation and event concentration. The two sequence methods agreed strongly with each other but only partly matched the four primary profiles. This suggests that the profiles combine several aspects of participation that are not fully reflected in week-by-week patterns alone. Dynamic time warping was not used because the sequences were already aligned by relative course progress; allowing additional shifts could hide meaningful differences in when activity occurred.

The South African setting extends the evidence base for learning analytics beyond the contexts most commonly represented in international research \cite{prinsloo2022}. Its importance lies not only in geographical coverage but also in showing the risks of comparing digital records across courses with different structures and participation conditions. Checking the influence of observation-window length and reporting assignment support provides a more transparent basis for educational profiling, although the study did not evaluate an operational intervention.

\section{Limitations and Future Research}
\label{sec:limitations}

The study used data from one university (2013--2019), did not capture offline participation and included only records with at least two active weeks; replication in contemporary and fully online settings is therefore needed. Duration adjustment reduced associations with observation-window length but did not account for course design, discipline, assessment timing or access conditions. Although the four-profile consensus was selected as the primary detailed solution, not every criterion favoured it; it depended partly on K-means and K-medoids, and Sustained active assignments were less certain. Agreement with sequence analyses was modest, indicating that week-by-week sequences reproduced only part of the detailed profiles. Future research should test the framework across institutions, account for learners appearing in multiple courses and course differences, and examine links with achievement and course design without treating the profiles as fixed risk labels.

\section{Conclusion}
\label{sec:conclusion}

This study introduced and evaluated a consensus framework that combines several clustering methods and forms of participation measures while accounting for observation-window length. Expressing the measures as rates or shares did not make records from shorter and longer courses comparable. Course-grouped adjustment substantially reduced this problem: profile membership accounted for 38.18\% of the variation in observation-window length before adjustment and 2.87\% after adjustment.

The results supported a broad two-profile grouping and a more detailed four-profile grouping: Gradually declining, Early-concentrated, Cyclical variable and Sustained active. Most records were consistently assigned, although uncertainty was greatest for Sustained active. The two week-by-week sequence analyses agreed strongly with each other but only modestly with the four-profile grouping. The profiles should therefore be understood as patterns that combine several aspects of participation, rather than as fixed learner types.

By checking the influence of observation-window length, comparing results across clustering methods and forms of participation measures, and reporting assignment consistency, the framework provides a more transparent and robust basis for profiling participation over time in higher education.

\begin{credits}

\subsubsection{\discintname}
The authors have no competing interests to declare that are relevant to the content of this article.
\end{credits}
%
%
%
%
%
%
\bibliographystyle{splncs04}
\bibliography{references}

\end{document}